%% file: main.tex
\documentclass[preprint,journal]{vgtc}            

\pdfoutput=1

\usepackage{stfloats}
\AtBeginDocument{%
  \captionsetup[figure]{labelsep=period}%
  \captionsetup[table]{labelsep=period}%
}

\onlineid{1702}

\vgtccategory{Application}

\title{Walking through Discussions: A Mobile Visual Analytics System for In-Situ Group Discussion Analysis}

\author{%
 \authororcid{Yiping Sun}{0009-0000-3986-7638}, Ziyao Kang, 
 \authororcid{Wei Zeng}{0000-0002-5600-8824}, 
 Minli Wu, and 
 \authororcid{Jiazhi Xia*}{0000-0003-4629-6268}
}

\authorfooter{
 \item
 	Yiping Sun, Ziyao Kang, Minli Wu, and Jiazhi Xia are with 
    the Central South University.
 	E-mail: \{yipingsun, kangzy, 8209220206, xiajiazhi\}@csu.edu.cn.
 \item
 	Wei Zeng is with the Hong Kong University of Science and Technology (Guangzhou) and the Hong Kong University of Science and Technology.
    E-mail: weizeng@hkust-gz.edu.cn.
 \item 
    Jiazhi Xia is the corresponding author.
}

\abstract{
  Group discussion-based teaching is widely used to foster collaborative learning, yet teachers in physical classrooms often struggle to simultaneously monitor multiple groups and quickly diagnose a target group before intervening. Existing visual analytics tools primarily support post-hoc analysis on desktop, providing limited support for in-situ walk-around teaching. To address this gap, we present \textit{MobileGroupVis}, a mobile visual analytics system for in-situ analysis of classroom group discussions. \textit{MobileGroupVis} integrates multi-group monitoring, single-group diagnosis, and instructional intervention into a concise analytical workflow tailored for small-screen touch interaction. The system is powered by a lightweight streaming analysis pipeline that converts group audio into structured discussion data and further extracts interaction patterns, topic progression, and topic deviation through a dialogue analysis module. To enable both glanceable overview and traceable diagnosis, we design six coordinated views, including a compact glyph that visually encodes word count, interaction intensity, and topic deviation for efficient cross-group comparison and anomaly localization, along with detailed views for opinion evolution, interaction dynamics, topic coverage, and dialogue records. We evaluate \textit{MobileGroupVis} through two case studies and expert interviews. The results provide preliminary evidence that \textit{MobileGroupVis} supports teachers in understanding discussion processes, identifying groups in need of attention, and facilitating in-class intervention.
}


\teaser{
  \centering
  \includegraphics[width=0.88\linewidth]{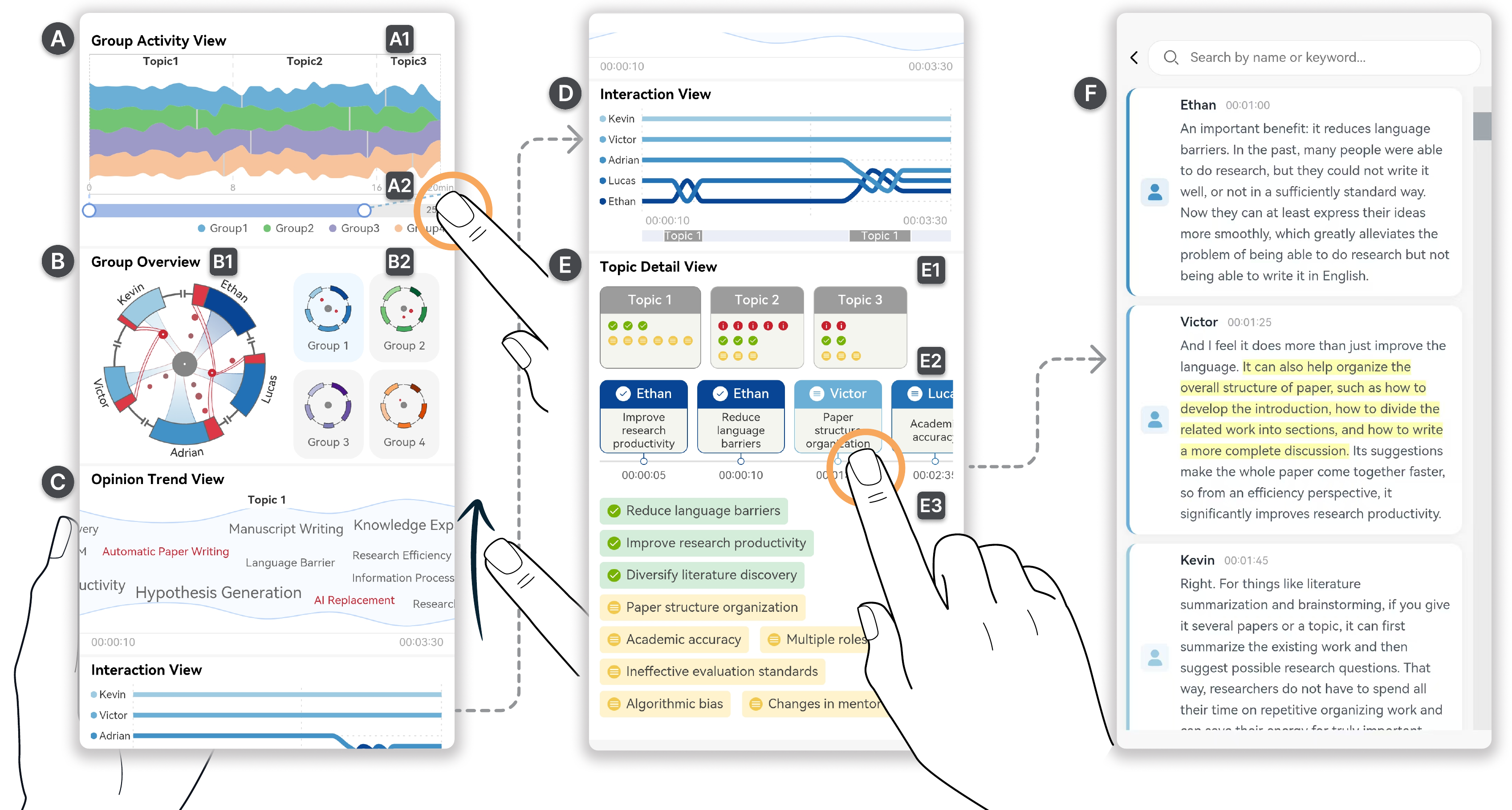}
  \vspace{-3mm}
  \captionsetup{labelsep=period}
  \caption{
  	\textbf{\textit{MobileGroupVis} interface}. \textit{Group Activity View} (A) presents the changes in overall discussion activity across groups. \textit{Group Overview} (B) uses a compact glyph to compare word count, interaction intensity, and topic deviation across multiple groups, helping teachers identify the group that requires priority attention. \textit{Opinion Trend View} (C) shows the opinion evolution within the anomalous group. \textit{Interaction View} (D) presents the interaction dynamics among group members. \textit{Topic Detail View} (E) provides information on the coverage of topics and opinions. \textit{Conversation Detail View} (F) displays the dialogue records of classroom discussions.
  }
  \vspace{-2mm}
  \label{fig:teaser}
}

\graphicspath{{figs/}{figures/}{pictures/}{images/}{./}} 

\usepackage{tabu}                      
\usepackage{booktabs}                  
\usepackage{lipsum}                    
\usepackage{mwe}                       
\usepackage{ccicons}                   

\usepackage{mathptmx}                  
\usepackage{algorithm}
\usepackage{algpseudocode}
\usepackage{amsmath}
\usepackage{url}

\begin{document}

\input{tex/1_introduction}
\input{tex/2_related_work}
\input{tex/3_observational_study}
\input{tex/4_system_overview}
\input{tex/5_data_processing}

\input{tex/6_MobileGroupVis}
\input{tex/7_evaluation}
\input{tex/8_discussion}
\input{tex/9_conclusion_and_future_work}

\section*{Supplemental Materials}
Our supplemental materials are available at \url{https://osf.io/r6x4k/overview?view_only=fb13281c978b43268df7b22c5c5eed32}. 
We provide the observational study materials, consent form template, preset topic contents for the two classroom cases, data processing details, the expert interview protocol, the baseline system interface, the source code, and a demonstration video of \textit{MobileGroupVis}.

\vspace{1mm}
\noindent
\textbf{CRediT authorship contribution statement}. Yiping Sun: Formal analysis, Investigation, Visualization (supporting), Writing – original draft. Ziyao Kang: Software (lead), Data curation, Validation. Wei Zeng: Methodology (supporting), Visualization (lead), Writing – review \& editing. Minli Wu: Software (supporting). Jiazhi Xia: Conceptualization, Methodology (lead), Writing – review \& editing, Funding acquisition, Project administration, Resources, Supervision.

\acknowledgments{
We would like to thank the domain experts and anonymous reviewers for their constructive comments. This paper is partially supported by National Natural Science Foundation of China (NO. U23A20313, 62372471) and The Key Project of Xiangjiang Laboratory (NO. 23XJ01011).
}

\bibliographystyle{abbrv-doi-hyperref}

\bibliography{reference}

\end{document}

%% file: tex/1_introduction.tex

\firstsection{Introduction}
\maketitle


Grounded in collaborative learning theories \cite{ally2004foundations,laal2012benefits,laal2012collaborative}, group discussion–based instruction is a pedagogical approach in which students engage in dialogue and collaboration around a specific topic under the guidance of an instructor, thereby promoting knowledge construction through communication and negotiation among peers \cite{peltoniemi2025understanding,van2019systematic}. 
Through in-depth interaction and collaborative problem solving within groups, this approach not only stimulates students’ self-directed inquiry and critical thinking, but also deepens their understanding of knowledge through the exchange of diverse peer perspectives \cite{smith2011combining}. 
Prior studies have demonstrated that this method can significantly improve learning performance \cite{smith2009peer}, and it has been broadly recognized in instructional reforms across multiple disciplines \cite{porter2011peer,vickrey2015based}.

Despite these benefits, effectively implementing group discussion in real classrooms remains challenging. 
In practice, multiple groups often engage in discussion simultaneously, and their discussion processes are dynamic, asynchronous, and highly contextual. 
Teachers need to continuously decide which group to attend to, when to intervene, and how to intervene. 
On the other hand, timely teacher intervention is critical to maintaining the quality of group discussion, and teachers’ expertise and experience play a key role in guiding both the direction and depth of students’ thinking \cite{doyle1977practicality}. 


 Our classroom observations and expert interviews (\cref{sec:obs_study}) indicate that, during in-situ walk-around teaching, teachers mainly face two challenges. First, they find it difficult to effectively monitor multiple groups at the same time, since teachers typically move back and forth among groups and can only obtain fragmented local observations.
 Second, teachers struggle to accurately assess the discussion status of problematic groups prior to intervention.
 When approaching a group, teachers often lack access to its recent discussion context and must ask students to recap, as video recordings or simple views cannot adequately support in-situ context reconstruction.
 Taken together, these challenges can be summarized into two core analytical tasks: 1) multi-group monitoring, and 2) rapid single-group diagnosis before intervention.


To support these analytical tasks, existing tools still fall short of meeting the needs of walk-around teaching in real classrooms. 
Although existing visual analytics systems for collaborative learning can reveal process information such as topic evolution, interaction relations, and student performance \cite{shi2018meetingvis,kui2025grouptrackvis,ngoon2024classinsight,jia2025high,chen2024stugptviz,Lianen2025Visual}, they are still primarily designed for post hoc analysis or desktop-based use. For instance, GroupTrackVis \cite{kui2025grouptrackvis} depends on multimodal data collected in group discussions, and its heavy data processing pipeline restricts the system to offline post-class analysis scenarios.
The observational study suggests that in-situ classroom use is constrained by fragmented attention, limited screen space, and imprecise touch interaction. 


To address these limitations, we present \textit{MobileGroupVis}, a mobile visual analytics system for in-situ analysis of classroom group discussions. 
We organize multi-group monitoring, single-group diagnosis, and instructional intervention into a short analytical path tailored to small-screen touch interaction and fragmented classroom attention. 
The system is driven by a lightweight streaming analysis pipeline (\cref{sec:data_process}) that transforms streaming group audio into structured discussion data through speech recognition and speaker identification, and further derives information such as interaction relations, topic progression, and topic deviation through a dialogue analysis module. 
We further design six coordinated views (\cref{sec:vis_design}) to support both multi-group monitoring and diagnosis of anomalies within a single group. 
In particular, we design a compact glyph for cross-group comparison that jointly encodes three key attributes: word count, interaction intensity, and topic deviation. 
\textcolor{black}{The design enables teachers to scan multiple group states within limited screen space and identify anomalous groups that may require priority attention.}
The remaining views characterize the discussion process of an anomalous group from different perspectives.
\textcolor{black}{The evaluation shows that \textit{MobileGroupVis} can effectively help teachers} monitor multiple groups and identify anomalous groups, and also trace their discussion processes, understand the causes of anomalies, and make more informed decisions for subsequent instructional intervention.


The contributions of this work are as follows:
\begin{itemize}
  \item 
    Through classroom observations and expert interviews, we identify the key bottlenecks of group discussion analysis in real-world walk-around teaching and summarize the design requirements and design criteria for in-situ mobile visual analytics.
  \item 
    We design and implement \textit{MobileGroupVis}, a mobile visual analytics system {\color{black} for in-situ monitoring, diagnosis, and intervention during walk-around teaching.} {The system} integrates {\color{black} a streaming pipeline that transforms live audio into structured data,} a compact glyph for cross-group comparison and anomaly localization, and a set of coordinated views for single-group diagnosis.
  \item 
    {\color{black}Through two case studies grounded in live classroom deployments and expert feedback, we evaluate how \textit{MobileGroupVis} supports walk-around teaching and distill design lessons for in-situ mobile learning analytics.}
\end{itemize}

%% file: tex/2_related_work.tex
\section{Related Work}

\subsection{Collaborative Learning Analytics}

Collaborative learning analytics employs computational methods to measure and model interaction data generated in collaborative learning, with the goal of revealing collaboration mechanisms and optimizing learning outcomes \cite{catasus2025collaborative}. Beyond learning outcomes alone, this line of research emphasizes a deeper understanding of complex collaborative behaviors and more precise intervention through the quantification of interaction processes. Traditional approaches to learning assessment have mainly focused on final group outcomes, paying limited attention to the interaction process itself and thus making it difficult to identify issues such as discourse domination or ineffective interaction \cite{dillenbourg1996evolution}. In recent years, the research focus has gradually shifted from static statistical indicators to the dynamic characteristics of collaborative processes \cite{wise2023nine}. Existing work on quantifying collaborative processes can be broadly divided into two directions: one focuses on the temporal modeling of social-cognitive interaction networks\cite{saqr2023temporal, vasco2024comparing, ouyang2023artificial}, while the other emphasizes multimodal behavior pattern recognition that extends attention to nonverbal cues\cite{chejara2023exploring,vrzakova2020focused,whitehead2025utilizing,LearnDialogue-Ma-LAK-2022,Xu2023Multimodal}.

However, existing studies primarily focus on post hoc offline analysis and rely on offline processing and desktop environments, which limits their ability to provide teachers with timely feedback in authentic classroom settings. 
In contrast, our work is designed for classroom walk-around scenarios and places greater emphasis on multi-group overviews and rapid diagnosis rather than in-depth analysis of a single group. 
Existing methods still remain insufficient in terms of streaming support and mobile adaptation.
\textcolor{black}{Following its shift from outcomes to interaction process, our dialogue analysis targets interaction relations, topic progression and deviation rather than final group outcomes.}

\subsection{Visual Analytics for Collaborative Learning}



The collaborative learning community has increasingly recognized the importance of integrating automated analysis with visualization\cite{chejara2024bringing,wang2025selfservice}.
A large body of prior work has focused on the in-depth exploration of textual discussion data in online learning environments, relying on asynchronous or semi-synchronous text-based media, such as forums, chatrooms, or interaction logs\cite{tan2021integrated, fu2016visual,fu2018visforum, fu2018t, chen2024stugptviz}. 
For example, iForum\cite{fu2016visual} investigates student interactions in MOOC forums. 
Through novel visual representations of posts, users, and threads, the system helps instructors comprehensively understand interaction patterns in forum discussions, including topic initiation, reply flows, and inter-group communication relationships. 
Based on long-term dialogue data between students and ChatGPT, StuGPTViz \cite{chen2024stugptviz} uses a customized Interaction Tree to depict the evolution of student–ChatGPT interaction patterns and compare behavioral differences across students.
Recent work has gradually begun to emphasize visual support for classroom discourse processes \cite{jia2025high, ngoon2024classinsight, kui2025grouptrackvis, keelawat2025dynamite,echeverria2024teamslides}. 
For instance, ClassInSight\cite{ngoon2024classinsight} supports teachers’ post hoc review and reflection on classroom discussions by mapping classroom speech data into multi-granularity visual forms, such as participation ratio charts and turn-taking trees. 


Overall, existing visualization work for collaborative learning either targets post hoc offline analysis and lacks support for streaming data, or is designed primarily for desktop environments, resulting in complex interfaces and cumbersome interactions. \textcolor{black}{There is limited support for the effective in-situ analysis of group discussions in classroom walk-around environments.} Our work addresses this gap with a mobile visual analytics system that is designed to assist teachers in conducting in-situ analysis in real classrooms.
\textcolor{black}{
Specifically, we design mobile-friendly views on a phone from live-streamed audio, enabling a seamless in-situ workflow for monitoring, diagnosis, and intervention.}

\subsection{Mobile Visualization}


Due to the shift toward ubiquitous data access, there is an increasing trend in developing mobile visualizations\cite{lee2021mobile, lee_2018_mobilevis}. 
Integrating visualization into portable devices involves not only the physical constraint of limited screen space \cite{kim2021data}, but also challenges related to fragmented attention and increased cognitive load in highly dynamic contexts \cite{grioui2024micro,While2024Glanceable}.
Efforts have been devoted to analyzing and comparing visual designs suitable for small screens \cite{brehmer2019visualizing, brehmer2020comparative,Blascheck2023PartToWhole,Cajamarca2023HealthVisChileanOlderAdults}. 
Other studies have drawn on the idea of responsive design while emphasizing the preservation of consistent data semantics across devices \cite{hoffswell2020techniques,kim2021design,kim2021automated,wu2020mobilevisfixer,zeng2023semi}.
Kim et al. \cite{kim2021automated} showed that when screen width is limited, dynamically adjusting the aggregation level of charts or modifying visual encodings can preserve the core distributional characteristics of data as much as possible within constrained space.
Beyond adapting desktop visualizations to smartphones, some studies have specifically designed visual analytics for mobile devices\cite{langner2021marvis, whitlock2019designing, zeng2023semi}. 
For instance, Whitlock et al. \cite{whitlock2019designing} explored bringing visual analytics directly into fieldwork settings to bridge the temporal gap between traditional data collection and subsequent analysis.
Portable mobile devices enabled rapid, high-level data overviews that helped field personnel quickly build global situational awareness, demonstrating the value of in-situ analysis in highly dynamic contexts. 


Mobile visualization aligns closely with teachers’ needs for walk-around monitoring and intervention in offline classroom settings, further underscoring the need to bring complex collaborative learning data to mobile platforms.
As such, we opt for a mobile visualization approach that prioritizes in-situ analysis and glanceable feedback.
\textcolor{black}{These mobile visualization principles shape our visual design, with cross-group glyph compressing each group's state into a compact form, and overview+detail to keep interaction light under fragmented attention.}

%% file: tex/3_observational_study.tex
\section{Observational Study}
\label{sec:obs_study}

\subsection{Experts’ Conventional Practice and Bottlenecks}


To gain an in-depth understanding of group discussion and teaching challenges, we collaborated closely with five domain experts (\textbf{E1}–\textbf{E5}) for three months.
All experts have more than five years of teaching experience and have long used group discussion as an instructional strategy.
\textbf{E1} and \textbf{E2} are senior professors in Computer Science and Technology with substantial experience in frontline teaching and education. 
\textbf{E3} is a professor specializing in visualization and visual analytics. 
\textbf{E4} is an educational data analyst. \textbf{E5} is a mobile development instructor. 
Except for \textbf{E3}, all other experts are independent of this study.


We observed five 45-minute in-person classes, each followed by a 30-minute semi-structured interview.
{\color{black}For every class, we recorded the audio and marked the teacher's movements and interventions.
Every interview was recorded and transcribed. 
Two authors then independently coded the transcripts and marks through thematic analysis and resolved disagreements through discussion.
The interview guide and coding scheme are provided in the supplemental material.}
All participants provided informed consent, and the observational study was approved by the Medical Ethics Committee of Central South University under CSUMEC-E2026025. The thematic analysis reveals that this approach of walking around the classroom exhibits several cognitive and behavioral bottlenecks:


\begin{enumerate}[label=\textbf{B\arabic*.}, leftmargin=*]
\item
\textbf{Attention conflicts and perceptual blind spots.}
Both \textbf{E1} and \textbf{E2} highlighted that teachers' inherently serial attention conflicts with simultaneous group discussions. As \textbf{E1} described, ``\emph{When I stop to listen to Group A, what is happening in Groups B and C on the other side of the room—whether an intense argument or complete silence—falls outside my perceptual field}.''  Relying strictly on physical senses restricts teachers to local, transient cues, hindering a global overview of classroom dynamics.


\item
\textbf{Context-deprived in-situ interventions.}
\textbf{E3} mentioned the diagnostic challenge of approaching stalled groups without historical context. Verbally querying students to recover recent discussions is inefficient and interrupts their cognitive flow. Furthermore, \textbf{E4} explained, ``\emph{Interventions made without a clear grasp of the group’s discussion trajectory may disrupt the rhythm and coherence of the ongoing discussion}.''


\item
\textbf{Desktop mobility constraints.} 
\textbf{E5} noted that podium computers are impractical during group discussion. He explained, ``\emph{If I have to return to the podium just to check the analytics, that breaks my physical connection with students}.'' Consequently, this spatial constraint forces a detrimental trade-off between accessing data insights and maintaining in-situ engagement.
\end{enumerate}

\begin{figure*}[t!]
    \centering
    \includegraphics[width=0.99\linewidth]{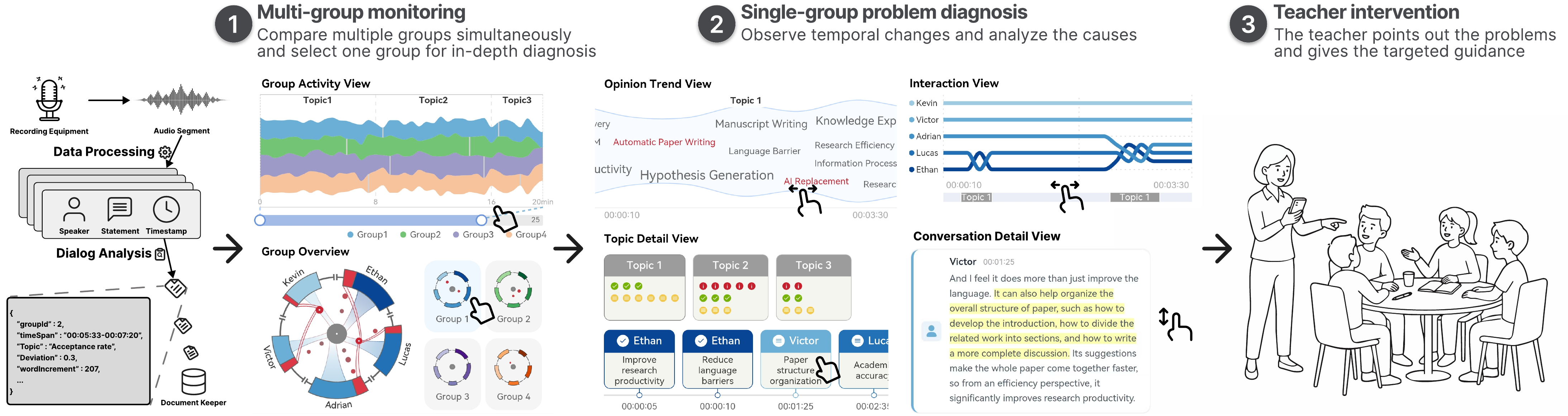}
    \vspace{-2mm}
    \caption{
    \textbf{Walking through discussion workflow for analyzing classroom group discussions with \textit{MobileGroupVis}}. 
    1) The teacher first observes and compares the overviews of all groups to identify the target group that requires priority attention. 
    2) \textcolor{black}{After selecting a target group, the teacher leverages multi-perspective diagnostic insights to infer possible causes of the anomalies.} 
    3) The teacher then combines these insights with on-site observations to carry out instructional intervention. 
    }
    \vspace{-3mm}
    \label{fig:workflow}
\end{figure*}

\subsection{Experts’ Needs and Expectations}


Drawing on these observed bottlenecks, we distilled five design requirements to support in-situ guidance via mobile visual analytics.



\begin{enumerate}[label=\textbf{R\arabic*.}, leftmargin=*]
\item
\textbf{Support lightweight in-situ analytics.}
\textbf{E1} and \textbf{E2} expressed a need to prioritize groups for intervention during walk-around teaching.
Since teachers split their attention moving around the classroom and interacting with students, the system should provide simple, mobile-friendly interactions that avoid complex menus. 
\textcolor{black}{Visual designs should be compact and surface useful information for classroom decisions while omitting details that do not inform teaching actions.} 
Furthermore, low-latency processing is crucial to synchronize insights with ongoing discussions.


\item
\textbf{Provide a cross-group overview of discussions.} 
The system should provide an overview tailored to the limited screen space of a mobile device. 
This macro-level view should intuitively visualize the discussion states of all groups, such as their current activity levels and stages of topic progression. 
As such, \textcolor{black}{teachers can maintain cross-group awareness} at a glance when walking around, allowing them to strategically optimize their monitoring paths.


\item
\textbf{Support rapid identification of target groups.} 
Given the severe constraints on teachers' cognitive bandwidth during classroom instruction, the system should proactively highlight potentially anomalous groups. 
Specifically, it should identify groups experiencing prolonged silence or those whose discourse exhibits substantial semantic deviation. 
Such a guide-oriented design can help teachers first allocate their limited attention to the groups that most need intervention.

\item
\textbf{Support rapid retrospective diagnosis of an individual group.} 
Once a target group has been identified, the system should provide drill-down analytical capabilities to support cause diagnosis. 
Teachers should be able to quickly inspect the detailed discussion process of that group on a mobile device, including opinion trend distributions, student interaction patterns, and summaries of discussed opinions. 
\textcolor{black}{ 
This approach mitigates the issue of missing context and supports evidence-informed guidance.}

\item
\textbf{Support traceable discussion records.} 
The system should provide traceable discussions that allow teachers to review the corresponding text segments of the discussion, the speakers involved, and the temporal order of utterances, while also marking topic changes with concise summaries. 
Based on such records, teachers can verify analytical results, understand the causes of problems, and support the formulation of targeted intervention strategies.

\end{enumerate}

%% file: tex/4_system_overview.tex
\section{System Overview}


To address these design requirements, we propose \textit{MobileGroupVis}, \textcolor{black}{a visual analytics system designed to help teachers detect and diagnose anomalous group discussions, and intervene in situ.} 
As illustrated in \cref{fig:workflow},
the workflow starts with a backend data preprocessing module and a dialogue analysis module. 
The data preprocessing module captures, segments, and cleans streaming classroom audio, utilizing speech recognition to output structured JSON transcripts. Subsequently, the dialogue analysis module performs lightweight streaming computations to continuously extract multidimensional features—encompassing activity metrics, interaction dynamics, topic deviation, and opinion trends. These incremental results are cached to drive updates and interactive exploration in the mobile frontend.

The analytical workflow forms a continuous macro-to-micro loop. Teachers begin in the \textit{Group Activity View} to identify cross-group topic transitions, potential anomalies, and activity shifts (\textbf{R1}, \textbf{R2}, \textbf{R3}). They then use the \textit{Group Overview} to compare discussion statuses and deviations, prioritizing a target group for diagnosis (\textbf{R1}, \textbf{R2}, \textbf{R3}). This selection triggers coordinated updates across three micro-views: the \textit{Opinion Trend View} reveals keyword trajectories and deviations (\textbf{R4}); the \textit{Interaction View} exposes participation balance, visualizing dominant speakers or interaction breakdowns (\textbf{R4}); and the \textit{Topic Detail View} tracks opinion coverage to guide interventions (\textbf{R4}, \textbf{R5}). Finally, teachers ground these insights in raw transcripts via the \textit{Conversation Detail View} \textbf{(R5}). Post-intervention, they return to the macro-views to monitor other groups, completing the iterative guidance loop.

%% file: tex/5_data_processing.tex
\section{Group Discussion Data Processing}
\label{sec:data_process}

\begin{figure}[t]
  \centering 
  \includegraphics[width=0.99\columnwidth]{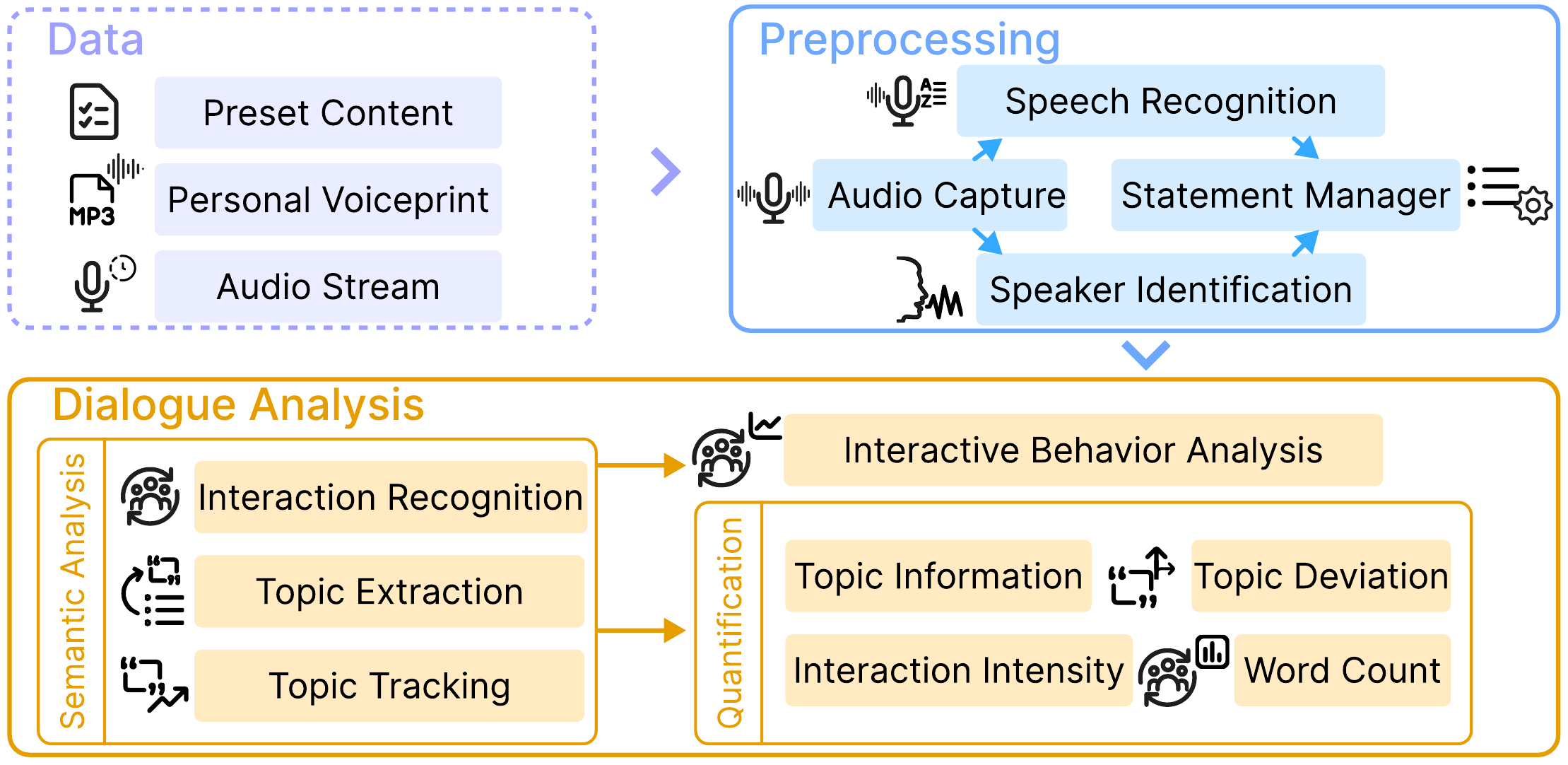}
  \vspace{-3mm}
  \caption{%
  	\textbf{The data analysis workflow}.
    It consists of two components:
    1) a data preprocessing module converts the collected audio into structured data; and 2) a dialogue analysis module uses lightweight streaming computation to extract and update information. 
  }
    \vspace{-4mm}
  \label{fig:Data Processing}
\end{figure}

This section introduces the data analysis workflow of the system,
as shown in \cref{fig:Data Processing}.

\subsection{Data and Preprocessing Module}

{\color{black}In the data preprocessing module, each group's streaming audio is segmented into 5-second windows to balance update responsiveness and semantic completeness, and non-speech intervals are filtered out.
We chose this window length because shorter segments often contain incomplete utterances, whereas longer segments increase the delay between classroom speech and visual updates.}
Valid speech segments are transcribed by an ASR API and matched against pre-registered student voiceprints for speaker identification using 1:N matching with confidence-based thresholding.
Since our work focuses on group discussion, the system registers only student voiceprints and does not register the teacher's voiceprint.
The resulting structured utterances, including timestamps, speakers, and text, are stored in the statement manager and periodically dispatched in batches to the dialogue analysis module for parallel processing.

\subsection{Dialogue Analysis Module}


After obtaining structured dialogue text, the dialogue analysis module performs interaction recognition, topic extraction, and topic tracking, followed by interaction behavior analysis and quantitative computation.

\subsubsection{Semantic Analysis}


Semantic analysis is implemented with the Qwen-Max model\cite{yang2024qwen25}, which supports interaction recognition, topic extraction, and topic tracking in a unified framework.
We choose LLMs because classroom discussion transcripts often preserve disfluencies, repeated fragments, and interrupted or incomplete utterances.
In such settings, LLMs are generally more robust than traditional topic analysis methods in semantic understanding and cross-sentence reasoning\cite{Long2024Evaluating,Suraworachet2024Predicting,Acosta2025Recognizing,yang2024qwen25}.


\vspace{1mm}
\noindent
1) \textit{Interaction Recognition}.
Interaction recognition determines whether an utterance responds to another member's speech.
Given the current utterance and recent dialogue context, the LLM identifies the reply target based on semantic relevance and outputs the corresponding speaker identifier or "None".
Each utterance is then assigned a reply-object field to construct the dialogue interaction structure.


\vspace{1mm}
\noindent
2) \textit{Topic Extraction}.
Topic extraction identifies the core content of discussion within each time window.
Given the current utterances, historical results, and the teacher's predefined topics, the LLM outputs a matched topic, related opinion point, and candidate keywords.
These outputs are standardized through topic ID validation, filtering of invalid opinion points, and keyword consistency checking.
\textcolor{black}{The result for each window, including its assigned topic, serves as input to topic tracking.}


\vspace{1mm}
\noindent
3) \textit{Topic Tracking}.
{\color{black}Topic tracking aims to maintain temporally coherent topic progression and detect topic switches.
The module leverages the extracted per-window topics from topic extraction, predefined topic order, and historical context to monitor topic progression over time.}
When another predefined topic becomes consistently more relevant across recent windows, a topic switch is recorded.
{\color{black}Topic matching occurs only once in topic extraction, while topic tracking handles only temporal sequencing and switch detection.}

\subsubsection{Interaction Behavior Analysis}


To identify interacting members within a specific period, the system analyzes interaction behaviors among group members.
We design a rule-based algorithm using topic consistency, temporal continuity, and reply relations.
\textcolor{black}{Specifically, the algorithm traverses time-ordered utterances with reply relations. The detailed procedure is provided in the supplemental material.}
If two adjacent utterances remain under the same topic, occur within a predefined time threshold, and satisfy the reply condition, they are assigned to the same interaction group; otherwise, a new group is created.
For each identified interaction group, the system records its start time, end time, associated topic, and the set of participating members.



\subsubsection{Quantitative Computation}


1) \textit{Topic Deviation Analysis}.
Topic deviation measures how much the current discussion content differs from the teacher's predefined topics and opinion points.
We
treat each discussion turn as the analysis unit, and the textual content of utterances is semantically encoded into a high-dimensional vector representation.
In implementation, we use the paraphrase-multilingual-MiniLM-L12-v2 model\footnote{\url{https://huggingface.co/sentence-transformers/paraphrase-multilingual-MiniLM-L12-v2}}.
Previous studies have shown that this class of multilingual embedding models achieves strong performance on semantic similarity and cross-lingual tasks \cite{wang2021minilmv2}.
We compute topic similarity using cosine similarity and define topic deviation as one minus this value.
A larger deviation indicates that the current discussion is less aligned with the predefined topic.

\vspace{1mm}
\noindent
2) \textit{Discussion Behavior Statistics}.
To characterize discussion activity and interaction behaviors, we compute word count-based statistics at both group and student levels, by summing the words across all utterances produced by a group within a time window, while student word count is computed from the utterances produced by each student.
Based on the interaction behavior analysis, utterances with identified reply relations are regarded as interactive utterances.
For each student, interaction intensity is computed by summing the word counts of the interactive utterances in which the student participates.
{\color{black}We use utterance-level word count as a basis for both discussion activity and interaction intensity.
For activity, it reflects verbal contribution over time.
For interaction intensity, it is applied only after reply relations are identified, measuring how much content is exchanged rather than how often turns occur.}
In addition, topic duration is computed as the time interval between the detected start and end times of each topic, reflecting the amount of discussion time allocated to different topics.

%% file: tex/6_MobileGroupVis.tex
\section{\textit{MobileGroupVis}}
\label{sec:vis_design}




The design of \textit{MobileGroupVis} follows the following design criteria:

\begin{enumerate}[label=\textbf{DC\arabic*.}, leftmargin=*]

\item
\textbf{Glanceable Visualization.}
To accommodate teachers' fragmented attention, the system \textcolor{black}{should support quick scanning for action-oriented classroom use.}
\textcolor{black}{Visualizations should be compact and show the information most relevant to teachers' in-class decisions rather than every available detail,} and maintain a consistent visual grammar to minimize cognitive load and support cross-group comparison.


\item
\textbf{Concise Analytical Workflow.}
Given imprecise touch interaction on mobile devices, the system requires a streamlined workflow.
The analytical workflow—multi-group monitoring, single-group diagnosis, and instructional intervention—must be executed with minimal swipes and taps.
Large touch targets, progressive drill-down, and shallow menus ensure teachers can access key evidence with minimal interaction.


\item
\textbf{Overview + Details.}
Analyzing discussions requires both cross-group comparisons and single-group inspections.
Therefore, the system must provide a global overview for efficient attention allocation, coupled with on-demand details to reveal possible causes and support intervention decisions.
Coordinated views and temporal alignment should ensure traceable exploration between these levels.


\end{enumerate}
\subsection{System Design}

\subsubsection{Group Activity View}

\begin{figure}[t]
  \centering 
  \includegraphics[width=0.95\columnwidth]{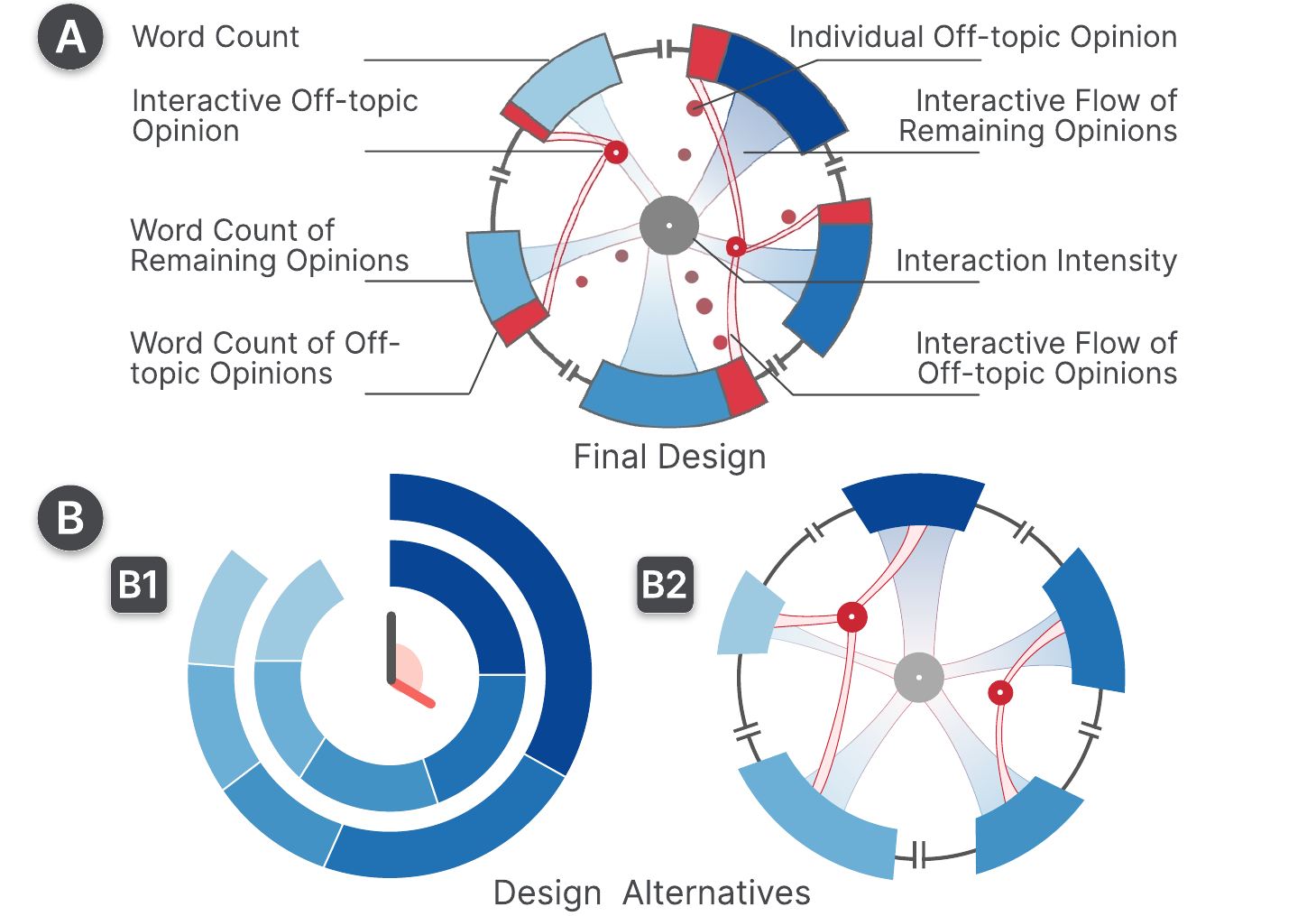}
  \vspace{-2mm}
  \caption{
  	\textbf{Glyph designs for Group Overview}.
  	The final glyph design (A) and two alternatives (B) present information such as group interaction behaviors and topic deviation.
  }
  \vspace{-3mm}
  \label{fig:Design Alternatives}
\end{figure}


\textit{Group Activity View} (\hyperref[fig:teaser]{\cref*{fig:teaser}(A)}) provides a temporal overview of all groups, supporting users in identifying activity shifts, topic progression, and anomalous periods (\textbf{R1}, \textbf{R2}, \textbf{R3}).

\vspace{1mm}
\noindent
\textbf{Visual Design.}
This view adopts a ThemeRiver~\cite{havre2000themeriver} to depict the continuous variation in group activity over time. 
Height of each river encodes word count of the corresponding group.
To align discussions with instructional tasks, gray dashed lines on the timeline mark scheduled periods, while white solid lines indicate actual topic transitions.
This enables users to assess engagement across topics, helping reveal overall participation declines or specific groups struggling to progress.

\vspace{1mm}
\noindent
\textbf{Interaction.}
We provide two temporal filtering mechanisms.
Users can swipe a range slider (\hyperref[fig:teaser]{\cref*{fig:teaser}(A2)}) at the bottom to isolate anomalous activities or salient cross-group differences.
Alternatively, tapping a `Topic' (\hyperref[fig:teaser]{\cref*{fig:teaser}(A1)}) above the ThemeRiver filters its scheduled discussion period.
Both interactions trigger coordinated updates across all views, maintaining temporal consistency for subsequent analysis.

\subsubsection{Group Overview}

\vspace{1mm}
\noindent
\textit{Group Overview} (\hyperref[fig:teaser]{\cref*{fig:teaser}(B)}) summarizes each group's discussion behavior (word count, interaction intensity) and content (topic deviation), supporting group comparisons and priority judgment (\textbf{R1}, \textbf{R2}, \textbf{R3}).

\vspace{1mm}
\noindent
\textbf{Visual Design.}
Following the overview + detail design criterion, this view employs a dual-granularity glyph design.
Small multiples of coarse-grained glyphs (\hyperref[fig:teaser]{\cref*{fig:teaser}(B2)}) summarize the status of all groups, while a fine-grained glyph (\hyperref[fig:teaser]{\cref*{fig:teaser}(B1)}) details the selected group for anomaly diagnosis.
Specifically, we designed a donut-based fine-grained glyph (\hyperref[fig:Design Alternatives]{\cref*{fig:Design Alternatives}(A)}) integrating students' word count, interaction intensity, and topic deviation.
The ring comprises student-specific segments, differentiated by monochromatic luminance.
Segment arc length encodes total word count, with a red sub-arc highlighting off-topic contributions.
Furthermore, high-contrast markers are overlaid to flag anomalous patterns at a glance.


The interior of the glyph encodes group interactions into three distinct categories: on-topic interaction, interactive off-topic opinion, and individual off-topic opinion.
{\color{black}First, for on-topic interaction, the radius of a central, gray hollow circle encodes the group-level interaction intensity.
This metric is computed by averaging the student-level interaction intensities within the group, and is subsequently normalized.}
Blue ribbons link student segments to this center, with their widths representing each student's on-topic interaction ratio.
Second, 
red hollow circles denote interactive off-topic opinions.
Their radius encodes intensity, distance from the center maps semantic deviation, and connecting ribbons identify contributors.
Third, individual off-topic opinions are represented by solid circles where the radius encodes word count.
To ensure rapid perceptual discrimination in dynamic settings, we adopt redundant visual encoding \cite{munzner2014visualization}, and their degree of deviation is encoded using both color and position.
Derived from the fine-grained design, the coarse-grained glyph retains the same three metrics: word count, interaction intensity, and topic deviation.
Based on expert feedback, we arrange these glyphs as small multiples to maximize the data-ink ratio and facilitate cross-group visual comparisons.

\vspace{1mm}
\noindent
\textbf{Design Alternatives.}
During the iterative design process, we considered two alternative glyphs (\hyperref[fig:Design Alternatives]{\cref*{fig:Design Alternatives}(B)}).
The first, a nested donut chart (\hyperref[fig:Design Alternatives]{\cref*{fig:Design Alternatives}(B1)}), encoded word count and interaction intensity on separate rings, using clock-like hands for topic deviation.
However, it only revealed aggregate group-level deviation without isolating individual off-topic instances.
The second, a standard donut with inner encodings (\hyperref[fig:Design Alternatives]{\cref*{fig:Design Alternatives}(B2)}), suffered from visual clutter due to overlapping ribbons' origins.
Our final design resolves this overlap issue while explicitly distinguishing between interactive and individual off-topic opinions.

\vspace{1mm}
\noindent
\textbf{Interaction.}
Tapping a right-side group facet updates the left focal area with its fine-grained glyph.
This highlighted selection triggers coordinated updates across all views, enabling rapid transitions between multi-group monitoring and single-group diagnosis.
Additionally, tapping red hollow circles or segments reveals keyword cues for interactive and individual off-topic opinions.

\subsubsection{Opinion Trend View}


Designed for rapid single-group diagnosis, \textit{Opinion Trend View} (\hyperref[fig:teaser]{\cref*{fig:teaser}(C)}) visualizes the temporal evolution of discussion content.
It highlights dominant opinions, off-topic opinions, and potential stagnation, enabling teachers to assess topic progression (\textbf{R4}).

\vspace{1mm}
\noindent
\textbf{Visual Design.}
We employ a temporally aligned word cloud where the x-axis represents time and the y-axis encodes word count.
Keywords are positioned based on their temporal occurrence.
Text size encodes mention frequency, while color indicates correctness (red for off-topic, gray for on-topic).
This layout allows users to track fleeting versus sustained opinions and monitor topical progression.
Unlike static keyword lists, this representation directly visualizes opinion evolution, facilitating a deeper process-oriented understanding.

\vspace{1mm}
\noindent
\textbf{Interaction.}
Users can swipe to pan across time periods or pinch to zoom for a broader temporal overview.
Tapping any keyword reveals its underlying transcript in \textit{Conversation Detail View}.

\subsubsection{Interaction View}


Designed for single-group diagnosis, \textit{Interaction View} (\hyperref[fig:teaser]{\cref*{fig:teaser}(D)}) visualizes intra-group communication patterns.
It enables teachers to assess participation balance, and identify dominant speakers or interaction breakdowns by revealing interaction dynamics over time (\textbf{R4}).

\vspace{1mm}
\noindent
\textbf{Visual Design.}
Inspired by a rope-twisting metaphor, we visualize abstract interactions through dynamic curves along a horizontal time axis, with students arranged vertically.
When interactions occur, their respective curves intertwine into a spiral structure, where the intertwined length encodes the interaction duration.
Consistent colors ensure stable cross-view mappings, while bottom-aligned topic labels contextualize these behaviors.
This compact design effectively illustrates temporal interaction sequences within limited mobile screen space.

\vspace{1mm}
\noindent
\textbf{Interaction.}
Users can swipe to pan across the timeline, enabling detailed inspection of interaction dynamics.

\subsubsection{Topic Detail View}



\textit{Topic Detail View} (\hyperref[fig:teaser]{\cref*{fig:teaser}(E)}) details intra-group opinions, enabling teachers to track topic progression, identify missing expectations, and discover emergent ideas (\textbf{R4}, \textbf{R5}).

\vspace{1mm}
\noindent
\textbf{Visual Design.}
This view integrates three components to inform intervention strategies: a topic card (\hyperref[fig:teaser]{\cref*{fig:teaser}(E1)}), an opinion card (\hyperref[fig:teaser]{\cref*{fig:teaser}(E2)}), and an opinion list (\hyperref[fig:teaser]{\cref*{fig:teaser}(E3)}).
The topic card summarizes opinion coverage (uncovered, discussed, and emergent) via icon counts, enabling rapid assessment of discussion completeness and omissions.
The opinion card chronologically aligns key opinions along a timeline, displaying the contributor and a brief summary.
The opinion list provides detailed context, using consistent color-coding to map actual opinions against predefined expectations.

\vspace{1mm}
\noindent
\textbf{Interaction.}
Tapping a topic card triggers updates in the opinion card and list.
Additionally, users can swipe to pan across historical opinions, or tap a specific opinion card to access its underlying transcript.

\subsubsection{Conversation Detail View}


Serving as the details-on-demand component, \textit{Conversation Detail View} (\hyperref[fig:teaser]{\cref*{fig:teaser}(F)}) presents raw transcripts, enabling teachers to ground their diagnostic insights in the original context (\textbf{R5}).

\vspace{1mm}
\noindent
\textbf{Visual Design.}
Chronologically ordered dialogue cards display the speaker, timestamp, and raw transcript.
A top search bar enables rapid retrieval by keyword or speaker.
Ultimately, this design closes the analytical loop, allowing teachers to drill down from high-level visual diagnoses to raw records to formulate targeted interventions.

\vspace{1mm}
\noindent
\textbf{Interaction.}
Users can scroll to navigate the transcripts or use keyword search to retrieve specific utterances.

\subsection{\color{black}Implementation}
{\color{black}

\textit{MobileGroupVis} adopts a client/server architecture.
The backend is implemented in Python with Flask for data processing and calls the Tencent Cloud ASR and speaker identification APIs.
The frontend is developed with HTML, CSS, and JavaScript, with D3.js \cite{hoque2020searching} used to implement the coordinated mobile visualizations.
The frontend communicates with the backend via RESTful APIs and is packaged as an Android application using HBuilderX.


Before deployment, teachers prepare a JSON configuration file that specifies the preset content required by the system, including group information, student voiceprint file names, topic order, topic descriptions, and expected opinions.
This JSON file, together with the student voiceprint files, is uploaded to the backend.
The system uses the JSON file as the core configuration file and registers the student voiceprints with the speaker-identification API by group.


During classroom deployment, each group has a microphone placed on its table to capture group-level discussion audio.
The audio streams are transmitted via Bluetooth to a laptop equipped with an Intel Core i7 processor and 32 GB of RAM, which runs the \textit{MobileGroupVis} backend.
The backend processes the incoming streams incrementally and sends the analysis results to the \textit{MobileGroupVis} mobile app through the local network.
As teachers walk around the classroom, they use the app to view cross-group states and inspect selected groups.
}

\begin{figure*}[t!]
    \centering
    \includegraphics[width=0.98\linewidth]{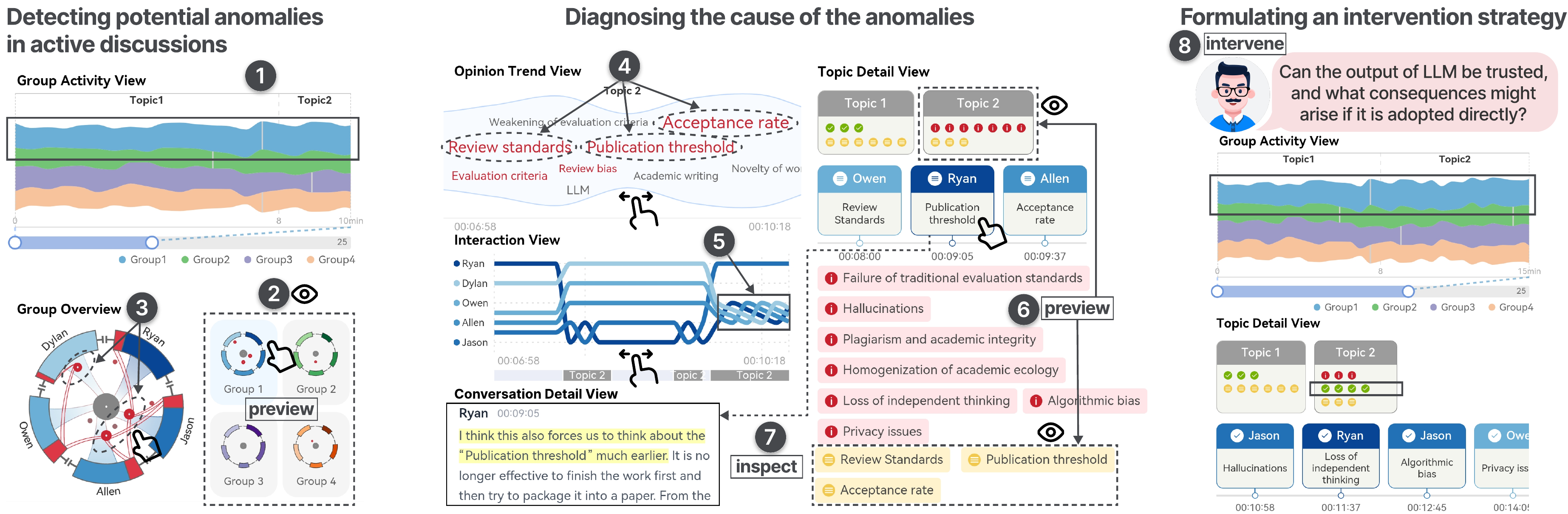}
    \vspace{-2mm}
    \caption{
    \textbf{Case I: Identifying and Redirecting Topic Deviation in Active Discussions.}
    Through cross-group comparison, \textbf{E1} (1)–(3) identified Group 1 as highly active yet showing significant topic deviation.
    He then (4)–(5) determined that the group was gradually deviating from the topic amid high interaction intensity, and (6)–(7) confirmed that several preset opinions had been omitted while the discussion continued to focus on non-target content.
    Finally, \textbf{E1} (8) posed a follow-up question on-site to guide the discussion back to the instructional goal.
    }
    \vspace{-3mm}
    \label{fig:case1}
\end{figure*}

%% file: tex/7_evaluation.tex
\section{evaluation}


We evaluate \textit{MobileGroupVis} through two live classroom deployments (\cref{sec:case_study}), and expert interviews with seven experts (\cref{sec:expert_interview}).

\subsection{Case Study}
\label{sec:case_study}




{\color{black}We conducted two pilot deployments in real courses, where students held group discussions for about 25 minutes while the expert used \textit{MobileGroupVis} on a mobile phone.
Before each session, the experts received a brief walkthrough of the system's visual encodings and interactions for about 25 minutes.}
{\color{black}
\textbf{E1} and \textbf{E3} conducted the experiments via live deployment.
We set up the experiments in real classrooms, with each setup lasting 15 minutes.
During each experiment, the microphones in each group captured the per-group audio, and the system logged its analysis results.
One co-author acted as a non-participant observer and took field notes on the expert's movement, system use, target-group selection, intervention timing, and intervention questions.
Two authors jointly reviewed the recordings, the system logs, and the field notes.
Next, \textbf{E2}, \textbf{E4}, and \textbf{E5} used \textit{MobileGroupVis} with the recorded audio streams from these two classes.
The recordings were fed into the same streaming analysis pipeline in chronological order. 
Their observations and feedback from \textbf{E1} and \textbf{E3} are reported in \cref{sec:expert_interview}.
}

{\color{black} Participants were informed of the recording procedure and
provided informed consent before data collection.
Audio, transcripts, and voiceprint files were used only for research, with access limited to the research team and the instructor.
The study was approved by the Medical Ethics Committee of Central South University under CSUMEC-E2026025.}

\subsubsection{Case I: Identifying and Redirecting Topic Deviation in Active Discussions}
\label{sssec:case_i}


We conducted a pilot deployment of the system in a course on academic writing and research ethics.
The class was divided into four groups, each consisting of five students.
The discussion topics focused on the benefits, risks, and ethical boundaries of using large language models (LLMs) in academic writing~\cite{kusumegi2025scientific,liang2025quantifying,porsdam2024guidelines}.
This case illustrates how \textit{MobileGroupVis} helped expert \textbf{E1} identify a group whose discussion appeared active on the surface but had substantively deviated from the topic during walk-around teaching, and how the system supported targeted in-situ intervention, as shown in \cref{fig:case1}.


\textbf{Detecting potential anomalies in active discussions.}
At the beginning of the discussion, students quickly became engaged, with members of each group sitting together and actively participating.
About 10 minutes into the discussion, \textbf{E1} walked from the back of the classroom toward the front.
Along the way, he first glanced at the \textit{Group Activity View} (\hyperref[fig:case1]{\cref*{fig:case1}(1)}) and found that while most groups showed relatively stable discussion rhythms during the current topic phase, the river for Group 1 was noticeably taller than those of the other groups, indicating that this group had maintained a consistently high word count.
The students in this group also appeared highly engaged: several students were speaking in rapid turns, with obvious gestures and energetic exchanges.
However, when \textbf{E1} further compared this group with the others in the \textit{Group Overview} (\hyperref[fig:case1]{\cref*{fig:case1}(2)}), he noticed that although Group 1 had both high word count and high interaction intensity, the red circles in the glyph were also highly prominent, especially several off-topic opinions generated through interaction (\hyperref[fig:case1]{\cref*{fig:case1}(3)}).
This pattern alerted \textbf{E1} that the group might not be engaged in high-quality in-depth discussion, but was instead collaboratively developing a line of discussion that had gradually deviated from the target topic.


\textbf{Diagnosing the possible cause of the anomaly.}
To verify this suspicion, \textbf{E1} walked toward the group while inspecting the detailed views.
First, in the \textit{Opinion Trend View} (\hyperref[fig:case1]{\cref*{fig:case1}(4)}), he observed that the group's earlier keywords had indeed centered on the benefits of LLMs.
However, over the most recent three minutes, a set of keywords weakly related to the current Topic 2 (the risks of LLMs) began to appear repeatedly and gradually occupied the center of the view.
\textbf{E1} then tapped the red circles for Group 1 in the \textit{Group Overview} to inspect the off-topic opinions, and the highlighted keywords included acceptance rate, publication threshold, and review standards.
He next examined the \textit{Interaction View}  (\hyperref[fig:case1]{\cref*{fig:case1}(5)}), which showed that the group's interaction was highly active during the current topic and mainly concentrated among four students.
This suggested that the current issue was that most members of the group were collectively driving the discussion away from the assigned topic through intensive interaction.
To further determine whether this topic deviation had already affected the group's progress on the task, \textbf{E1} checked the \textit{Topic Detail View} (\hyperref[fig:case1]{\cref*{fig:case1}(6)}).
This view showed that the group had not yet covered the predefined opinions, with seven key opinions still undiscussed.
Meanwhile, three newly emerged opinions were not fully aligned with the goal of the current topic.
\textbf{E1} then looked at the discussion records in the \textit{Conversation Detail View} (\hyperref[fig:case1]{\cref*{fig:case1}(7)}) and confirmed that these four students kept elaborating on and questioning these three new opinions. 


\textbf{Formulating an intervention strategy.}
Based on this diagnosis, \textbf{E1} carried out an immediate intervention (\hyperref[fig:case1]{\cref*{fig:case1}(8)}).
After approaching the group, he asked a follow-up question, ``\emph{Can the output of LLMs be trusted, and what consequences might arise if it is adopted directly}?'' 
This question required the students to connect their discussions with the key opinions that had not yet been covered. 
In this way, the intervention redirected the discussion back toward the instructional objective.
After intervening, \textbf{E1} continued walking along the classroom aisle and checked the system from time to time.
After that, Group 1 remained highly active, its keywords gradually became aligned with the topic, and the \textit{Topic Detail View} showed that the group had covered four predefined opinions under Topic 2.



\subsubsection{Case II: Addressing Discussion Stagnation and Participation Imbalance}
\label{sssec:case_ii}

\begin{figure*}[t!]
    \centering
    \includegraphics[width=0.99\linewidth]{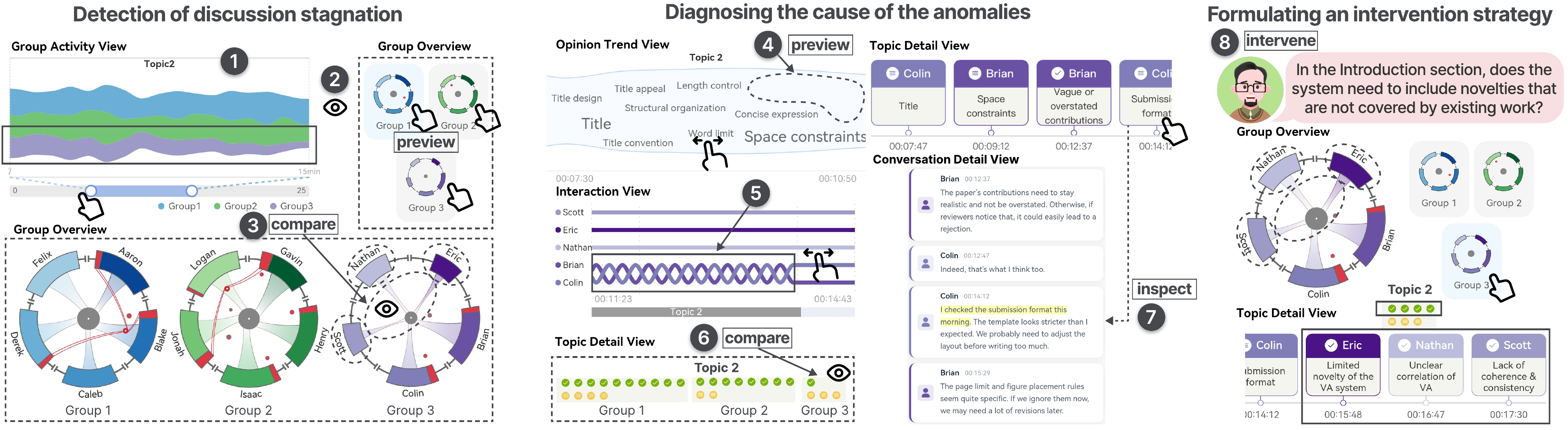}
    \vspace{-2mm}
    \caption{
    \textbf{Case II: Addressing Discussion Stagnation and Participation Imbalance.}
    Through cross-group comparison, \textbf{E3} (1)–(3) identified that Group 3 showed insufficient word count and interaction intensity, as well as imbalanced participation among members, suggesting that the discussion might be in a stagnant state.
    He then (4)–(5) determined that the group lacked opinion progression and exhibited anomalous interaction patterns, and (6)–(7), by comparing it with other groups, confirmed that the topic was progressing slowly and that interaction occurred only between two members.
    Finally, \textbf{E3} (8) invited the group members to respond to questions and introduced new directions for thinking.
    }
    \vspace{-3mm}
    \label{fig:case2}
\end{figure*}

We conducted another pilot deployment of the system in a visualization and visual analytics course.
The class was divided into three groups, each consisting of five students.
The discussion topic focused on common categories of criticism in visual analytics papers \cite{wu2022defence}.
This case illustrates how \textit{MobileGroupVis} helped expert \textbf{E3} identify a group exhibiting discussion stagnation and participation imbalance, and how the system supported targeted in-situ intervention, as shown in \cref{fig:case2}.


\textbf{Detection of discussion stagnation.}
At the beginning of the discussion, students actively engaged in the task.
About 15 minutes into the discussion, \textbf{E3} was patrolling along the classroom aisle.
At this point, he quickly glanced at the \textit{Group Activity View} (\hyperref[fig:case2]{\cref*{fig:case2}(1)}) and found that 
all groups had already been discussing Topic 2 for some time, while 
their activity levels had begun to diverge noticeably: Group 1 and Group 2 still maintained stable speaking patterns, whereas Group 3 showed a sustained downward trend in activity.
To further inspect Group 3's situation under Topic 2, \textbf{E3} dragged the left handle to the start time of Topic 2 to adjust the temporal window.
He then examined the cross-group differences in the \textit{Group Overview} (\hyperref[fig:case2]{\cref*{fig:case2}(2)}) within this time window and found that Group 3 not only had relatively low overall word count, but also showed substantially weaker interaction intensity than the other groups.
\textbf{E3} inferred that the group might be in a stagnant state and decided to move closer for further observation.


\textbf{Diagnosing the possible cause of the anomaly.}
When \textbf{E3} approached Group 3, 
he 
continued his assessment by combining on-site observation with the interface.
He noticed that two students in the group were looking down at their materials, while the other three students were not communicating.
\textbf{E3} further compared the fine-grained glyphs (\hyperref[fig:case2]{\cref*{fig:case2}(3)}) of the three groups.
The results showed that the word count and interaction intensity in Group 3 were mainly concentrated in two members (Colin and Brian), while the other members had shorter arc lengths and fewer interaction flows.
\textbf{E3} then examined the \textit{Opinion Trend View} (\hyperref[fig:case2]{\cref*{fig:case2}(4)}) to understand why the group had become stuck.
The keywords showed that the group had briefly discussed title and length limitation at the beginning of Topic 2, but over the following several minutes, the discussion centered on only a few overly general terms and failed to advance toward new opinions.
In the \textit{Interaction View} (\hyperref[fig:case2]{\cref*{fig:case2}(5)}), a twisted spiral segment indicated sustained interaction between two members, whereas the other three members remained outside the interaction, forming an anomalous interaction pattern.
\textbf{E3} further examined the \textit{Topic Detail View} (\hyperref[fig:case2]{\cref*{fig:case2}(6)}) and found that, although the group had already discussed one preset opinion and three newly emerged opinions, it still missed a considerable number of key opinions compared with the other groups.
He further examined the dialogue records in the \textit{Conversation Detail View} (\hyperref[fig:case2]{\cref*{fig:case2}(7)}), confirming that the group was relatively silent, the topic discussion had stalled, and no opinions were being further developed.


\textbf{Formulating an intervention strategy.}
After diagnosing the possible causes of discussion stagnation and participation imbalance, \textbf{E3} decided to intervene.
Rather than directly providing an answer, he first posed a question targeting the missing opinions to the three less-involved students, ``\emph{In the Introduction section, do we need to explain why the research problem should be addressed using visual analytics methods? Does the system need to include novelties that are not covered by existing work}?'' (\hyperref[fig:case2]{\cref*{fig:case2}(8)}).
Introducing new directions for thinking, \textbf{E3} helped the group break out of its repetitive loop.
About three minutes after the intervention, \textbf{E3} continued patrolling and revisited the system.
The group's total word count in the \textit{Group Activity View} gradually increased, and the distributions of speaking and interaction in the \textit{Group Overview} became more balanced.
The \textit{Topic Detail View} further showed that, after the intervention, the group successfully discussed three predefined opinions.
\textbf{E3} then returned to monitoring the other groups and continued making cross-group comparisons based on the overview while walking around the classroom.



\subsubsection{\textcolor{black}{Comparative Study}}
\label{ssec:comparative_study}

{\color{black} 
We conducted a comparison of \textit{MobileGroupVis} with a baseline system  on anomaly identification using the two classroom settings.
The baseline system simulated conventional classroom observation, where teachers could choose which group discussion to listen to and mark anomalies.
The study involved \textbf{E1} and \textbf{E3}, and to reduce memory effects, they cross-validated each other's settings using the baseline system: \textbf{E1} examined Case II, and \textbf{E3} examined Case I.}

{\color{black}
We used expert-annotated anomalies as ground truth, covering topic deviation, missing viewpoints, stagnation, imbalanced participation, and insufficient interaction.
Case I and Case II contained 10 and 11 anomalies, respectively. 
A detection was counted as correct if the teacher marked the same group and anomaly type within a one-minute window around the ground-truth timestamp.
With the baseline system, \textbf{E3} identified only three anomalies in Case I and \textbf{E1} identified four in Case II.
In comparison, \textit{MobileGroupVis} helped \textbf{E1} identify eight anomalies in Case I and helped \textbf{E3} identify nine for Case II.
In addition, \textit{MobileGroupVis} enables teachers to inspect cross-group summaries and diagnostic evidence, thereby substantially reducing overlooked anomalous behaviors.
Interaction logs further indicate that teachers mainly relied on diagnostic views during anomaly identification.
The \textit{Topic Detail View} was used most frequently, with 6.68 interactions per minute, followed by the \textit{Opinion Trend View} (4.48), \textit{Interaction View} (4.36), and \textit{Group Overview} (4.34).
This suggests that teachers did not simply scan group-level summaries, but actively inspected topic coverage, opinion evolution, and interaction patterns to understand why a group was anomalous.
}


\subsection{Expert Feedback}
\label{sec:expert_interview}



We collected expert feedback on the perceived usefulness, interpretability, usability, and deployment considerations of \textit{MobileGroupVis}.
We interviewed the five experts from \cref{sec:obs_study} (\textbf{E1}–\textbf{E5}) and two new experts (\textbf{P1} and \textbf{P2}). 
\textbf{P1} and \textbf{P2} have taught programming language and machine learning courses, respectively.



\noindent
\textbf{Procedure.}
{\color{black}The experts had distinct types of exposure to the system.
\textbf{E1} and \textbf{E3} gave feedback from their live classroom use in Case I and Case II, while \textbf{E2, E4}, and \textbf{E5} used the system with recorded audio streams from these sessions.}
For \textbf{P1} and \textbf{P2}, we first introduced the project background, including the analytical tasks and data sources, then demonstrated the system workflow and visual encodings using the audio stream from Case II in \Cref{sec:case_study}.
Next, they were asked to freely explore the system using a think-aloud protocol.
Considering that the new experts were unfamiliar with visual analytics, we adopted a co-discovery approach to help them learn how to use \textit{MobileGroupVis}.
Finally, we conducted a 30-minute semi-structured interview with each expert.
During this process, we recorded their comments and observations, {\color{black}and distinguished comments grounded in live classroom use from those based on streaming replay or demonstration-based exploration.
We reviewed the interview notes and think-aloud comments, grouped recurring feedback by theme, and summarized the results in three categories as follows.}


\noindent
\textbf{System Workflow.}
Most experts generally perceived the overall analytical workflow of \textit{MobileGroupVis} to be highly consistent with their classroom teaching practices.
\textcolor{black}{Based on in-situ experience, \textbf{E1} and \textbf{E3}} appreciated the way the system organizes analysis into a process of monitoring, diagnosis, and intervention, as it aligns with how teachers make intervention decisions under time pressure.
\textcolor{black}{Other experts also considered this workflow understandable and pedagogically meaningful.}
\textbf{E2} stated, ``\emph{The system accelerates the diagnostic process from first noticing a group's problem to understanding its specific details.
Compared with traditional classroom patrolling, it reduces the burden of repeatedly reconstructing context}.''
\textbf{P1} pointed out that the system would be especially valuable for maintaining global awareness during walk-around teaching and for diagnosing the possible causes of anomalies in a specific group before intervening.
\textbf{E4} and \textbf{E5} both emphasized that a short analytical path would be particularly important for classroom use.
As \textbf{E1} remarked, ``\emph{In real classrooms, what I care more about is whether I can quickly reach evidence that supports intervention, rather than performing complex, in-depth exploration}.''
\textbf{E1} and \textbf{E3} jointly stressed the importance of the traceability of raw discussion records.
The system enables them to validate high-level analytical insights against raw discussion snippets prior to intervention.
\textbf{E3} commented, ``\emph{Locating detailed dialogue records helped me build trust in the system and reduced my concern about interrupting students' discussion based on insufficient judgment}.''


\noindent
\textbf{Visualization and Interaction.}
All experts recognized the value of the mobile visualization.
\textbf{E1} and \textbf{E3} found \textit{MobileGroupVis} better suited to walk-around teaching than desktop-based analytical tools, as it allowed them to remain physically close to students.
As \textbf{E3} noted, ``\emph{Analyzing while walking with a phone in hand not only reduces the pressure of classroom guidance, but also saves the time I previously had to spend listening in on groups}.''
\textcolor{black}{Other experts also perceived the mobile format as appropriate for classroom teaching.}
The experts generally considered the combination of the \textit{Group Activity View} and \textit{Group Overview} to be highly valuable for rapid cross-group comparison.
\textbf{E4} observed that these two views could help teachers allocate attention more strategically across groups, rather than reacting only to nearby ones.
\textbf{E1} and \textbf{E2} further noted that the \textit{Group Overview} reveals whether the more urgent issue is inactivity, participation imbalance, or topic deviation.
As \textbf{E1} stated, ``\emph{This capability is especially important in classes where multiple groups discuss in parallel}.''
\textbf{P1} and \textbf{P2} were impressed by the visualization and interaction design.
\textbf{P2} commented, ``\emph{The glyph designs are visually appealing, and their large touch targets are easy to tap}.''
\textbf{P1} noted, ``\emph{The word cloud in the \textit{Opinion Trend View} constantly reminds me of the opinions currently under discussion, and the \textit{Topic Detail View} imposes almost no learning burden and is very easy to use}.''
Experts also acknowledged the value of the \textit{Interaction View}.
\textbf{E4} remarked, ``\emph{I like this intertwining spiral structure—it is aesthetically pleasing while allowing me to observe who is interacting, making it easier to identify interaction patterns}.''


\noindent
\textbf{Suggestions.}
The experts also proposed several suggestions for improvement.
\textbf{E3} hoped that the \textit{Opinion Trend View} could provide richer interactions.
He stated, ``\emph{Some keywords are quite small and require careful inspection to distinguish.
I suggest adding zooming interactions to reduce misclicks}.''
\textbf{E1} suggested adding an intervention review mode, so that teachers could relate the moment of intervention to the subsequent discussion outcomes and assess the effectiveness of instructional intervention.
\textbf{E5} suggested that the system provide more support for intervention itself.
He explained, ``\emph{For less experienced teachers, it would be more helpful if the system could further provide lightweight intervention cues, such as possible follow-up questions}.''
{\color{black}Some experts also noted that \textit{MobileGroupVis} should support, rather than replace, direct teacher-student communication.
While the system helps teachers recover discussion context, asking students to recount their discussions remains valuable for understanding their reasoning and supporting reflection.}
These suggestions point to promising directions for the future extension of \textit{MobileGroupVis}.

%% file: tex/8_discussion.tex
\section{Discussion}


This work extends the application of visual analytics for group discussions from post hoc review to in-situ classroom analysis, and illustrates the potential value of mobile visual analytics in educational settings. 
\textit{MobileGroupVis} focuses on teachers’ real-world practices of moving, observing, and making decisions during class. 
Therefore, the compact visual encodings and concise analysis workflow emphasized in \textit{MobileGroupVis} can be understood not only as interface design choices, but also as direct responses to the practical demands of instructional decision-making in classroom contexts.


\vspace{1mm}
\noindent
\textbf{Lessons Learned.} Through the design of \textit{MobileGroupVis}, we derived two main lessons. 
\textcolor{black}{First, walk-around teaching scenario prioritizes early anomaly identification over exhaustive analysis.} 
Since teachers often have limited time and attention for complex operations during class, a short analysis path and an at-a-glance overview are more important than complete information coverage. 
For example, small multiples provide an effective way to encode overview information for multiple groups.
{\color{black}
On the other hand, oversimplified design, such as a few computed indicators, may be easy to read and sufficient for lightweight monitoring. 
However, such design provides limited support for diagnosing why a group is anomalous or how teachers should intervene. 
Our design philosophy does not aim for a balance between information richness and cognitive load. 
Instead, we strive to raise the upper bound of this trade-off: compact glyphs keep group states glanceable, while lightweight touch interactions reveal denser diagnostic views only when a teacher selects a target group.
This makes details available on demand rather than imposed continuously.}

Second, analysis results should be verifiable. 
Conclusions alone are often insufficient for intervention, whereas discussion records with contextual backtracking better support teachers in understanding unusual situations.
{\color{black}
Merely relying on LLM-based analysis to determine intervention timing is not optimal. During our system development, we observed that misidentifications generally arise under distinct scenarios: overlapping or incomplete speech degrading ASR or speaker identification, short or shorthand utterances, gradual transitions between semantically close topics, or implicit replies. Consequently, topic switches may be detected late, replies may be misattributed, or emergent opinions may be misclassified as deviations. This can cause an on-track group to appear off-track, or a struggling group to go unnoticed.
\textit{MobileGroupVis} helps teachers cross-reference these analytical indicators against raw transcripts and real-time in-class observations prior to intervention.
By integrating visual analytics with an LLM-based pipeline, this framework is capable of elevating the system’s overall reliability and efficacy. Furthermore, future iterations will benefit from advances in LLMs, speech recognition, and speaker identification models.}


\vspace{1mm}
\noindent
\textbf{Generalization and Scalability.} Although \textit{MobileGroupVis} was designed for small-group discussion in classrooms, its workflow may be useful in other settings that require monitoring multiple discussions and intervention, such as seminars or collaborative training sessions. 
At present, the system performs well in small-scale classrooms with no more than 30 students, {\color{black}with an end-to-end update delay of approximately 15 seconds, which mainly came from speech recognition, speaker identification, and LLM-based semantic analysis.}
However, as classroom size increases, several scalability challenges may arise.
When the number of groups grows further, the burden of {\color{black} audio processing and} cross-group comparison on mobile devices increases for teachers.
\textcolor{black}{
More parallel audio streams and longer discussions would increase the load of speaker identification, LLM reasoning, and embedding computation. 
Meanwhile, more groups would make the cross-group glyphs denser, while longer discussions would make temporal views and transcripts harder to inspect. 
A larger number of students within each group would also make participation and interaction patterns more complex in the \textit{Group Overview} and \textit{Interaction View}. 
}

\vspace{1mm}
\noindent
\textbf{Limitations.} 
Our work still has several limitations. 
First, the current audio processing uses fixed 5-second audio segmentation to support streaming analysis.
While this balances update responsiveness and semantic completeness, it may truncate utterances or produce semantically incomplete inputs. 
{\color{black}Future work could explore adaptive segmentation based on utterance boundaries or dialogue turns.} 
Second, the analysis window length involves a trade-off: longer windows provide richer context but increase latency, whereas shorter windows improve responsiveness at the cost of analysis quality.
In addition, system performance depends heavily on upstream data quality. 
In real classrooms, noise, overlapping speech, and incomplete spoken expressions can reduce analysis accuracy. 
Although additional prompting can mitigate some of these issues, it also increases LLM reasoning time and latency.
Nevertheless, although the system is not fully real-time in a strict sense, it remains responsive enough for in-situ classroom support.
{\color{black}Last but not least, \textit{MobileGroupVis} records, transcribes, and visualizes student discussions, which may raise ethical and privacy considerations. 
In our deployments, students were informed of the recording procedure and provided consent before data collection. 
Still, students may feel uncomfortable or change their participation when they know their utterances can be inspected. 
Thus, we treat the system as only a means to support discussion rather than grading or surveillance evaluation.}

%% file: tex/9_conclusion_and_future_work.tex
\section{conclusion and future work}


We presented \textit{MobileGroupVis}, a mobile visual analytics system for the in-situ analysis of classroom group discussions.
To address teachers’ practical needs in walk-around teaching, we organized multi-group monitoring, single-group diagnosis, and instructional intervention into a short analytical path, and designed six coordinated views to help teachers better understand discussion states, identify anomalous groups, and diagnose the possible causes of anomalies. 
\textcolor{black}{Case studies and expert interviews provide preliminary evidence that \textit{MobileGroupVis} supported teachers in detecting discussion stagnation and participation imbalance, and in informing intervention decisions.} 
In future work, the system could incorporate richer multimodal data to enhance the understanding of classroom collaborative processes. It could also be extended to support larger-scale classrooms and be evaluated through broader user studies across diverse teaching scenarios. We envision that \textit{MobileGroupVis} can provide a new research entry point for the in-situ analysis of classroom group discussions and further promote the application of mobile visual analytics in authentic educational settings.